%% file: main.tex
\documentclass{tlp}
\usepackage{url}
\usepackage{graphicx}
\usepackage{subfigure}
\usepackage[dvips,all]{xy}
\usepackage{tikz}
\usetikzlibrary{shapes.geometric} 
\usetikzlibrary{shadows}

\newcommand\boundedtermsize{bounded term-size}

\newtheorem{theorem}{Theorem} % [section]
\newtheorem{example}{Example} % [section]

\newcommand{\algorithmprob}{PITA(PROB)}
\newcommand{\algorithmprism}{PITA(IND,EXC)}
\newcommand{\algorithmposs}{PITA(POSS)}
\newcommand{\algorithmvitind}{PITAVIT(IND)}
\newcommand{\comment}[1]{}

\begin{document}

\long\def\comment#1{}

\title[The PITA System]{The PITA System: Tabling and Answer Subsumption for Reasoning under Uncertainty}
\author[F. Riguzzi and T. Swift]
{FABRIZIO RIGUZZI \\
ENDIF -- University of Ferrara\\
Via Saragat 1, I-44122, Ferrara, Italy  \\
E-mail: fabrizio.riguzzi@unife.it
\and TERRANCE SWIFT \\
CENTRIA -- Universidade Nova de Lisboa \\
E-mail: tswift@cs.suysb.edu
}

\pagerange{\pageref{firstpage}--\pageref{lastpage}}
%\volume{\textbf{10} (3):}
%\jdate{March 2002}
\setcounter{page}{1}
%\pubyear{2002}
\maketitle

\label{firstpage}

\noindent
{\bf Note}: This article has been published in Theory and Practice of Logic Programming, 11(4--5), 433--449, 
\copyright Cambridge University Press, \\
%\href{http://dx.doi.org/10.1017/S147106841100010X}{
doi:10.1017/S147106841100010X\\
The caption of Figure \ref{biomine}, mistaken in the journal version, has been corrected.
\input{abstract}

\begin{keywords}
Probabilistic Logic Programming, Possibilistic Logic Programming, Tabling, Answer Subsumption, Program Transformation
\end{keywords}

\section{Introduction}
\label{intro}
Uncertainty, imprecision and vagueness are very important for modeling real world domains where facts can often not be ascertained with complete confidence. In the field of Logic Programming, there  have recently been many efforts to include these characteristics,
%in the language, 
originating whole research fields such as Probabilistic Logic Programming (PLP),  Possibilistic Logic Programming and Fuzzy Logic Programming.
In all three fields many approaches have been proposed for modeling uncertainty, imprecision and vagueness, obtaining new languages that are often equipped with efficient inference algorithms.

In Probabilistic Logic Programming, a large number of languages have
been independently proposed. Many of these however follow a common
approach, the distribution semantics \cite{DBLP:conf/iclp/Sato95}, and
in fact there are transformations for converting a program in one PLP
language into another PLP language \cite{VenVer03-TR,DeR-NIPS08}.
Examples of such PLP languages are Probabilistic Logic Programs
\cite{DBLP:conf/lpar/Dantsin91}, Probabilistic Horn Abduction (PHA)
\cite{DBLP:journals/ngc/Poole93}, Independent Choice Logic (ICL)
\cite{Poo97-ArtInt-IJ}, PRISM \cite{DBLP:conf/iclp/Sato95}, Logic
Programs with Annotated Disjunctions (LPADs),
\cite{VenVer04-ICLP04-IC} and ProbLog
\cite{DBLP:conf/ijcai/RaedtKT07}. 
% FR1: I think the difference between the languages and the systems must be stressed more. PHA is the only language that inposes restrictions on the programs, the other don't. E.g., the PRISM semantics is defined also for programs without the assumptions and is equivalent to LPADs, ProbLog and ICL. For PRISM, see the paper sato kameya JAIR2001 at https://ds.ing.unife.it/svn/problp/papers/related  It is the system that imposes the restriction for ease of evaluation.
%
Most of these languages impose few restrictions on the type of
programs they can evaluate -- ICL, LPADs and others for instance, have
been defined on normal programs with function symbols.  Accordingly,
we term systems that evaluate large classes of PLP programs {\em
  general} PLP systems.  However a great deal of efficiency and
scalability can be obtained by restricting how different explanations
are constructed and combined.  Such an approach is adopted
by the PRISM system \cite{PRISM-manual} which we refer to as a {\em
  restricted} PLP system.  Both general and restricted PLP systems
have advantages in different domains depending on the form of
uncertainty to be represented, the form of programs needed to model
problems, and on the scale of the problems to be solved.  

% FR: Probabilistic LPs are general, PHA  requires the mutual exclusion of rule bodies
% Does ProbLog or ICL allow negation?  
% FR ProbLog allows a restricted form of negation but it is going to allow full negation
% What is the complexity
% difference between PRISM and the others?  If there is a complexity
% difference, how is this affected in the size of translation between,
% say LPADs and PRISM?
% FR The PRISM semantics does not have restriction, it is the PRISM system that requires
% the independence and mutual exclusion conditions

\comment{
Previous
 While these languages share the
same distribution semantics, they have differences in the classes of
programs they can evaluate and in the ensuing complexity of
evaluation.  ICL, LPADs and ProbLog impose few restrictions on the
type of programs they can evaluate -- LPADs, for instance have been
defined on normal programs with function symbols.  As a result, we
term these approaches {\em general} PLP languages.  PRISM on the other
hand makes strong assumptions about the exclusiveness of clauses for a
predicate and about the independence of literals in a clause; as a
result it can combine the probabilities of different explanations of a
subgoal in a simpler manner than the general approaches.  We term
PRISM a {\em restricted} PLP language.  This restriction can be a
strength, as PRISM can be more scalable than the general approaches on
many problems.
}

Possibilistic Logic Programming models uncertainty by means of
possibility theory rather than probability theory. Possibilistic Logic
Programming aims at computing the degree of uncertainty of a query in
the form of a necessity measure. Given a possibilistic knowledge base,
inference rules have been developed for answering
queries~\cite{DBLP:journals/fss/DuboisP04a}.

In this paper we show that an inference technique and system developed
for general PLP called {\em Probabilistic Inference with Tabling and
  Answer subsumption (PITA)}, can be parameterized to efficiently
reason with different measures of uncertainty.  PITA translates a
general PLP program into a normal program that is evaluated by a
Prolog engine with tabling. The transformation adds an extra argument
to each subgoal to provide access to an auxiliary data structure used
in computing the uncertainty of the subgoal.  The transformed program
is evaluated using tabling to memo intermediate results and to support
well-founded negation, along with a tabling feature named {\em answer
  subsumption} to combine explanations 
%for goals 
from different
clauses, and a set of library predicates to interface with the
auxiliary data structure.

PITA was first presented in \cite{RigSwi10-ICLP10-IC} and addressed
general PLP using Binary Decision Diagrams (BDDs) as auxiliary data
structures.  That version of PITA, termed here \algorithmprob{}, was
compared with ProbLog, \texttt{cplint} \cite{Rig-AIIA07-IC} and CVE \cite{MeeStrBlo08-ILP09-IC}
%in \cite{RigSwi10-ICLP10-IC} 
and found to be fast and scalable.  In this paper we first consider a
parameterization called \algorithmprism{} and compare to the
restricted PLP system PRISM, one of the first and most widely used
systems for PLP.
%PRISM, which is one of the first and most used
%systems for probabilistic inference. PRISM makes some assumptions on
%the form of programs that simplify considerably the computation. In
%this paper we show that PITA can be slightly modified to exploit these
%assumption as well, obtaining an algorithm that turned out to be
Preliminary results show that \algorithmprism{} turns out to be faster than PRISM on complex queries
to a naive encoding of a Hidden Markov Model (HMM). When the optimized encoding proposed by
\cite{DBLP:conf/iclp/ChristiansenG09} is used, the timing result depend on the input data, with PRISM faster on random sequences and \algorithmprism{} faster on repeated sequences. When adapting PITA to compute
the most probable explanation of the query (or Viterbi's path), we
obtain similar performances in relation to PRISM.

Moreover, we show that PITA can be also be parameterized to \algorithmposs{}
to compute the necessity of formulas from Possibilistic Logic
Programs, and show the resulting implementation to be highly scalable.
%PITA's transformation and library predicates had to be
%modified in minor ways for implementing possibilistic inference
%rules. 
Together, these results show the versatility of the PITA algorithm,
and how the implementation can be easily adapted to support different
types of uncertain reasoning.

The paper is organized as follows. Section \ref{problp} presents
Probabilistic Logic Programming while Section \ref{posslp} discusses
Possibilistic Logic Programming. Section \ref{tabling} reviews tabling
and answer subsumption; while Section \ref{algorithm} presents the
PITA program transformation and \algorithmprob{}. 
%Section \ref{gpc} presents
%PITA for general probabilistic programs. 
In Section \ref{prism} we
describe \algorithmprism{}
% had to be modified for exploiting PRISM assumption
together with experimental results on an HMM dataset.
%, while in Section
%\ref{cg} we presents the experimental results for the optimized HMM
%program.  Section \ref{viterbi} shows that PITA can be used also for
%computing the Viterbi path. 
Section \ref{appposslp} presents
\algorithmposs{} for computing necessity levels from possibilistic programs.
%and Section \ref{conc} concludes the paper.

\section{Probabilistic Logic Programming}
\label{problp}
%In the field of Logic Programming, many languages follow the distribution semantics \cite{DBLP:conf/iclp/Sato95} or a variant thereof.
Various languages have been proposed in the field of Probabilistic Logic Programming, such as for example  Bayesian Logic Programs \cite{KerDeR00-ILP00-IC}, CLP(BN) \cite{SanPagQaz03-UAI-IC}
 or P-log \cite{DBLP:journals/tplp/BaralGR09}.
A large group of languages follows the distribution semantics \cite{DBLP:conf/iclp/Sato95} or a variant thereof.
In the distribution semantics a probabilistic logic program defines a probability distribution over a set of normal logic programs (called \emph{worlds}).
The distribution is extended to a joint distribution over worlds and queries and the probability of a query is obtained from this distribution by marginalization.

The languages differ in the way they define the distribution over logic programs. Each language allows probabilistic choices among atoms in clauses: 
Probabilistic Logic Programs, PHA, ICL, PRISM, and ProbLog allow probability distributions over facts, while LPADs allow probability distribution over the heads of  clauses.
%
% maybe mention the size complexity of the transformation along
% with modularity properties?  FR: the transformations have linear
% complexity (added), what bo you mean by modularity properties?
% I dont see how they have the same expressive power as you give
% examples below of programs that can't be evaluated by PRISM.
% FR1: they can't be evaluated by the PRISM system, but their translation into PRISM is semantically well defined 
%
%TLS: is this true for P-Log?
All these languages have the same expressive power: there are transformations with linear complexity that can convert each one into the others \cite{VenVer03-TR,DeR-NIPS08}. In this paper we will use LPADs because their syntax is the most general.

\begin{example} \label{markov-model}
The following LPAD $T_1$ captures a Markov model of length two with three states of which state 3 is an end state
$$
 \begin{array}{llll}
 C_1=&s(0,1):1/3 \vee s(0,2):1/3 \vee s(0,3):1/3. \\
 C_2=&s(1,1):1/3 \vee  s(1,2):1/3 \vee  s(1,3):1/3 &\leftarrow & s(0,1).\\
 C_3=&s(1,1):0.2 \vee  s(1,2):0.2 \vee  s(1,3):0.6 &\leftarrow & s(0,2).
 \end{array}$$
The predicate $s(T,S)$ models the fact that the system is in state $S$ at time $T$. Clause $C_1$ selects the first state, while clauses $C_2$
and $C_3$ select the second state depending on the value of the first. As state 3 is the end state, if $s(0,3)$ is selected at time 0, no state follows.
\end{example}
LPADs are sets of disjunctive clauses in which each atom in the head
is annotated with a probability. If the probabilities in the head do
not sum up to 1, an extra dummy atom $null$ is implicitly assumed to
represent the remaining probability mass and is such that it does not
appear in the body of any clause. A ground LPAD clause
represents a probabilistic choice among the normal program clauses
obtained by selecting one of the heads.

We now define the distribution semantics for the case in which a
program does not contain function symbols so that its Herbrand base is
finite \footnote{However, the distribution semantics for programs with
  function symbols has been defined as well
  \cite{DBLP:conf/iclp/Sato95,DBLP:journals/jlp/Poole00,RigSwi10-CILC10-NC}.}.
Let us first introduce some terminology.
An \emph{atomic choice} is a selection of the $i$\--th atom for a grounding $C\theta$ of a probabilistic clause $C$
and is represented by the triple $(C,\theta,i)$.

For example, $(C_2,\{\},1)$ is an atomic choice selecting atom $s(1,1)$ from  $C_2$
obtaining the clause
$$ \begin{array}{llll}
s(1,1)\leftarrow s(0,1).
 \end{array}$$
 A set of atomic choices $\kappa$ is \emph{consistent} if $(C,\theta,i)\in\kappa,(C,\theta,j)\in \kappa\Rightarrow i=j$, i.e., only one head is selected for a ground clause. 
For example $\kappa=\{(C_2,\{\},1),(C_2,\{\},2)\}$ is not consistent.
 
 A  \emph{composite choice} $\kappa$ is a consistent set of atomic choices.
The probability of composite choice $\kappa$  is 
$$P(\kappa)=\prod_{(C,\theta,i)\in \kappa}P_0(C,i)$$
where $P_0(C,i)$ is the probability annotation of head $i$ of clause $C$.
A \emph{selection} $\sigma$ is a total composite choice (one atomic choice
for every grounding of each probabilistic statement/clause).
For example, $\sigma=\{(C_1,\{\},1),(C_2,\{\},1),(C_3,\{\},2)\}$ is a selection for $T_1$.
A selection $\sigma$ identifies a  logic program $w_\sigma$ called  a \emph{world}.
The probability of $w_\sigma$ is $P(w_\sigma)=P(\sigma)=\prod_{(C,\theta,i)\in \sigma}P_0(C,i)$.
Since the program does not have function symbols the set of worlds is finite: $W_T=\{w_1,\ldots,w_m\}$ and  $P(w)$ is a distribution over worlds:  $\sum_{w\in W_T}P(w)=1$

\iffalse
Let the Herbrand base be $H_T=\{A_1,\ldots,A_n\}$. 
An Herbrand interpretation can be seen as an assignment of truth values to ground atoms, which, in probabilistic logic programming, are Boolean random variables. Following Bayesian network terminology, we indicate with $a_i$ the assignment of a truth value to $A_i$, so an interpretation can be represented as $I=\{a_1,\ldots,a_n\}$.

We can define the conditional probability of an interpretation given a world in the intuitive way:
$P(I|w)=1$ if $I$ if a model of $w$ and 0 otherwise.
In this way, we can have a joint distribution over interpretations and worlds:
$$P(I,w)=P(I|w)P(w)$$
By marginalizing over worlds we get the probability of an interpretation
$$P(I)=\sum_{w}P(I,w)=\sum_{w}P(I|w)P(w)=\sum_{w,\mbox{$I$ model of $w$}}P(w)$$
The distribution over interpretations can be seen as a joint distribution $P(A_1,\ldots,$ $A_n)$ over the atoms of $H_T$.
If the query is a ground atom $A_j$ of which we want to know the probability that is true
$P(A_j=true)=P(a_j)$, we can obtain it by marginalizing over the other atoms
$$P(a_j)=\sum_{a_i, i\neq j}P(a_1,\ldots,a_m)=\sum_{I, a_j \in I}P(I)$$
Alternatively, we can define the conditional probability of a query given a world:
$P(a_j|w)=1$ if $A_j$ is true in $w$ and 0 otherwise.
The probability of the query can then be obtained by marginalizing over the worlds
$$P(a_j)=\sum_{w}P(a_j,w)=\sum_{w}P(a_j|w)P(w)=\sum_{w\models A_j}P(w)$$
\fi
We can define the conditional probability of a query $Q$ given a world:
$P(Q|w)=1$ if $Q$ is true in $w$ and 0 otherwise.
The probability of the query can then be obtained by marginalizing over the query
$$P(Q)=\sum_{w}P(Q,w)=\sum_{w}P(Q|w)P(w)=\sum_{w\models Q}P(w)$$
%
%
% am I wrong, or is it more precisely \#P?
% FR the reference mentions that it is NP-hard
Inference in probabilistic logic programming is performed by finding
explanations for queries. An explanation is a composite choice such
that the query is true in all the worlds that are compatible with the
composite choice. The query is true if one of the explanations
happens, so the query is true if the disjunction of the explanations
is true, where each explanation is interpreted as the conjunction of
all its atomic choices. Each of these choices is associated to a
probability so the problem of computing the probability of the query
is reduced to the problem of computing the probability of a DNF
formula, which is an NP\--hard problem
\cite{DBLP:conf/iclp/KimmigCRDR08}. The most efficient way to date of solving
the problem makes use of Binary Decision Diagram (BDDs) that are used to
represent the DNF formula in a way that allows to compute the
probability with a simple dynamic programming algorithm
\cite{DBLP:conf/ijcai/RaedtKT07,Rig-AIIA07-IC,DBLP:conf/iclp/KimmigCRDR08,Rig08-ICLP08-IC,Rig09-LJIGPL-IJ,RigSwi10-CILC10-NC,RigSwi10-ICLP10-IC,Rig10-FI-IJ}.

\section{Possibilistic Logic Programming}
\label{posslp}

Possibilistic Logic \cite{DubLanPra-poss-94} is a logic of uncertainty that allows reasoning under incomplete evidence. In this logic,  the degree of necessity of a
formula expresses to what extent the available evidence entails the truth of the formula and  the
degree of possibility expresses to what extent the truth of the formula is not incompatible with
the available evidence.

Given a formula $\phi$, we indicate with
 $\Pi(\phi)$ its degree of possibility  and with 
$N(\phi)$ its degree of necessity. Their relation is established by
$N(\phi)=1-\Pi(\neg\phi)$.

A possibilistic clause is a first order logic clause $C$ to which a number is attached taken as a lower bound of its necessity or possibility degree. We consider here the possibilistic logic CPL1  \cite{DBLP:conf/iclp/DuboisLP91} in which only lower bounds on necessity 
are considered. Thus $(C,\alpha)$ means that $N(C)\geq \alpha$.
A possibilistic theory is a set of possibilistic clauses.

A possibility measure satisfies a possibilistic clause $(C,\alpha)$ if $N(C)\geq \alpha$ or equivalently if $\Pi(\neg C)\leq 1- \alpha$. A possibility measure satisfies a possibilistic theory if it satisfies every clause in it.
A possibilistic clause $(C, \alpha)$ is a consequence of a possibilistic theory $F$ if every possibility measure satisfying $F$ also satisfies $(C, \alpha)$.

Inference rules of classical logic have been extended to rules in possibilistic logic. Here we report  two sound inference rules \cite{DBLP:journals/fss/DuboisP04a}:
  \begin{itemize}
 \item $(\phi,\alpha),(\psi,\beta) \vdash (R(\phi,\psi),\min(\alpha,\beta))$ where $R(\phi,\psi)$ is the resolvent of $\phi$ and $\psi$ (extension of resolution)
 \item $(\phi,\alpha),(\phi,\beta) \vdash (\phi,\max(\alpha,\beta))$ (weight fusion)
\end{itemize}
A Possibilistic Logic Programming language has been proposed in \cite{DBLP:conf/iclp/DuboisLP91}. 
A Possibilistic Logic Program is a set of formulas of the form $(C,\alpha)$
 where $C$ is a definite program clause
% Mentioned in text below: 
%\footnote{Semantics for possibilistic normal programs has been given in \cite{DBLP:conf/micai/OsorioN09} and \cite{DBLP:journals/amai/NicolasGSL06}}
 $$H\leftarrow B_1,\ldots,B_n.$$
and $\alpha$ is a possibility or necessity degree. We consider the subset of this language that is included in CPL1, i.e., $\alpha$ is a real number in (0,1] that is a lower bound on the necessity degree of $C$.
The problem of inference in this language consists in computing the maximum value of $\alpha$ such that $N(Q)\geq \alpha$ holds for a query $Q$.
The above inference rules are complete for this language.

\begin{example} \label{poss-path}
The following possibilistic program computes the least unsure path in a graph, i.e., the path with maximal weight, the weight of a path being the weight of its weakest edge \cite{DBLP:conf/iclp/DuboisLP91}.
$$
 \begin{array}{llll}
(path(X,X),&&1)\\
(path(X,Y)\leftarrow path(X,Z),edge(Z,Y),&&1)\\
(edge(a,b),&&0.3)\\
\ldots
 \end{array}$$
\end{example}
We restrict our discussion here to positive programs.  However we note
that approaches for normal Possibilistic Logic programs have been
proposed in
\cite{DBLP:conf/lpnmr/NievesOC07,DBLP:journals/amai/NicolasGSL06,DBLP:conf/micai/OsorioN09} and
\cite{BSCV10}.

\input{tab}
\section{PITA for General Probabilistic Logic Programming}
\label{algorithm}
\paragraph*{The PITA Transformation.}
PITA computes the probability of a query from a probabilistic program
in the form of an LPAD by first transforming the LPAD into a normal
program containing calls to manipulate uncertainty information.  The
idea is to add an extra argument to each literal to access a data
structure containing the information that is necessary for computing
the probability of the subgoal.  The extra arguments of these literals
are combined using a set of general library functions:
\begin{itemize}
  \item \textit{init, end}: initialize and terminate the extra data structures necessary for manipulating uncertainty information
  \item \textit{zero(-D), one(-D), and(+D1,+D2,-DO), or(+D1,+D2, -DO), not(+D1,-DO)}: Boolean operations between uncertainty information data structures;
  \item \textit{add\_var(+N\_Val,+Probs,-Var)}: addition of a new multi-valued random variable with \textit{N\_Val} values and list of probabilities \textit{Probs};
  \item \textit{equality(+Var,+Value,-D)}: \textit{D} is a data structure representing \textit{Var=Value}, i.e. that the random variable {\em Var} is assigned {\em Value} in $D$;
  \item \textit{ret\_prob(+D,-P)}: returns the probability of the data structure \textit{D}.
  \end{itemize}
%
% also V -> Var to make get_var_n paramaters like those of var/3.
%
% TLS: Par -> Probs so add_var like get_var_n
%
%\textit{add\_var(+N\_Val,+Probs,-Var)} adds a new random variable associated to a new  instantiation of a rule with \textit{N\_Val} head atoms and parameters list \textit{Probs}.
The auxiliary predicate \textit{get\_var\_n(+R,+S,+Probs,-Var)} is
used to wrap \textit{add\_var/3} to avoid adding a new random variable
when one already exists for a given clause instantiation.  As shown
below, a new fact \textit{var(R,S,Var)} is asserted each time a new
random variable is created: 
%, where \textit{R} is an identifier for the LPAD clause,
%\textit{S} is a list of constants representing the clause's
%instantiation, and 
\textit{Var} is an integer that identifies the
random variable associated with clause \textit{R} under the grounding
represented by \textit{S}. 
{\em get\_var\_n/4} has the following definition

\[\begin{array}{ll}
get\_var\_n(R,S,Probs,Var)\leftarrow\\
\ \ \   (var(R,S,Var)\rightarrow true;\\
\ \ \    length(Probs,L), add\_var(L,Probs,Var), assert(var(R,S,Var))).
\end{array}\]
%where \textit{Probs} is a list of floats that stores the parameters in the head of rule \textit{R}.
% \textit{R}, \textit{S} and \textit{Probs} are input arguments while \textit{Var} is an output argument.
%
The PITA transformation applies to clauses, literals and atoms.  The
transformation for a head atom $H$, $PITA_H(H)$, is $H$ with the
variable $D$ added as the last argument.  Similarly, the
transformation for a body atom $A_j$, $PITA_B(A_j)$, is $A_j$ with the
variable $D_j$ added as the last argument.
%In either case for an atom $A$, $D(PITA(A))$ is the value of the last
%argument of $PITA(A)$, 
%
The transformation for a negative body literal $L_j=\neg A_j$, $PITA_B(L_j)$, is the Prolog conditional 
\[(PITA_B'(A_j)\rightarrow not(DN_j,D_j);one(D_j)),\]
 where $PITA_B'(A_j)$ is $A_j$ with the variable $DN_j$ added as the last argument.  
In other words, the input data structure, $DN_j$, is negated if it exists; otherwise the data structure for the constant function $1$ is returned.

\iffalse
 A non-disjunctive fact $C_r=H$ is transformed into the clause  

\[PITA(C_r)=PITA_H(H)\leftarrow one(BDD).\]

\noindent A disjunctive fact $C_r=H_1:\alpha_1\vee \ldots\vee H_n:\alpha_n$.
where the parameters sum to 1, is transformed into the set of clauses $PITA(C_r)$\footnote{The second argument of $get\_var\_n$ is the empty list because a fact does not contain variables since the program is \boundedtermsize{}.}

\[\begin{array}{ll}
PITA(C_r,1)=PITA_H(H_1)\leftarrow &get\_var\_n(r,[],[\alpha_1,\ldots,\alpha_n],Var),\\
&equality(Var,1,D).\\
\ldots\\
PITA(C_r,n)=PITA_H(H_n)\leftarrow &get\_var\_n(r,[],[\alpha_1,\ldots,\alpha_n],Var),\\
&equality(Var,n,D).\\
\end{array}\]

\noindent
In the case where the parameters do not sum to one, the clause is first transformed into
$H_1:\alpha_1\vee \ldots\vee H_n:\alpha_n\vee null:1-\sum_{1}^n\alpha_i.$
and then into the clauses above, where the list of parameters is $[\alpha_1,\ldots,\alpha_n,1-\sum_{1}^n\alpha_i]$ but the $(n+1)$-th clause (the one for $null$)
is not generated. 

\noindent
The definite clause $C_r=H\leftarrow L_1,\ldots,L_m$.
is transformed into the clause 

\[\begin{array}{ll} 
PITA(C_r) = PITA_H(H)\leftarrow &one(DD_0),\\
&PITA_B(L_1),and(DD_0,D_1,DD_1),\\
&\ldots,\\
&PITA_B(L_m),and(DD_{m},D_m,D).
\end{array}\]

\noindent
\fi
The disjunctive clause

\[C_r=H_1:\alpha_1\vee \ldots\vee H_n:\alpha_n\leftarrow L_1,\ldots,L_m.\]

\noindent
where the parameters sum to 1, is transformed into the set of clauses $PITA(C_r)$

\[\begin{array}{ll} 
PITA(C_r,1)=PITA_H(H_1)\leftarrow&one(DD_0),\\
&PITA_B(L_1),and(DD_0,D_1,DD_1),\ldots,\\
&PITA_B(L_m),and(DD_{m-1},D_m,DD_m),\\
&get\_var\_n(r,VC,[\alpha_1,\ldots,\alpha_n],Var),\\
&equality(Var,1,DD),and(DD_m,DD,D).\\
\ldots\\
PITA(C_r,n)=PITA_H(H_n)\leftarrow&one(DD_0),\\
&PITA_B(L_1),and(DD_0,D_1,DD_1),\ldots,\\
&PITA_B(L_m),and(DD_{m-1},D_m,DD_m),\\
&get\_var\_n(r,VC,[\alpha_1,\ldots,\alpha_n],Var),\\
&equality(Var,n,DD),and(DD_m,DD,D).
\end{array}\]

\noindent
where  $VC$ is a list containing each variable appearing in $C_r$.
%
\iffalse
If the parameters do not sum to 1, the same technique used for disjunctive facts is used.
\fi
\begin{example}
\label{ex-tra}
Clause $C_1$ from the LPAD of Example \ref{markov-model} is translated into
\[
 \begin{array}{llll}
 s(0,1,D)&\leftarrow& one(DD_0), get\_var\_n(1,[],[1/3,1/3,1/3],Var),\\
 &&equality(Var,1,DD),and(DD_0,DD,D).\\
 s(0,2,D)&\leftarrow& one(DD_0), get\_var\_n(2,[],[1/3,1/3,1/3],Var),\\
 &&equality(Var,1,DD),and(DD_0,DD,D).\\
 s(0,3,D)&\leftarrow& one(DD_0), get\_var\_n(3,[],[1/3,1/3,1/3],Var),\\
 &&equality(Var,1,DD),and(DD_0,DD,D).\\
\end{array}
\]

\end{example}
%TLS: Changed prob to genl-prob as we now have ind/exc as well.
In order to answer queries, the goal {\em genl\_prob(Goal,P)} is used, which is defined by 

\[
 \begin{array}{llll}
genl\_prob(Goal,P)&\leftarrow   init, retractall(var(\_,\_,\_)), \\
&add\_d\_arg(Goal,D,GoalD),\\
\ \ \   &(call(GoalD)\rightarrow ret\_prob(D,P) ; P=0.0),\\
&end.
\end{array}
\]

\noindent
where $add\_d\_arg(Goal,D,GoalD)$ implements $PITA_H(Goal)$.

\paragraph{Evaluating the Transformed Program.}
Various predicates of the transformed program should be declared as
tabled. For a predicate $p/n$, the declaration is
{\em table p(\_1,...,\_n,or/3-zero/1)},
which indicates that answer subsumption is used to form the disjunct
of multiple explanations. At a minimum, the predicate of the goal and
all the predicates appearing in negative literals should be tabled
with answer subsumption.  
%As shown in Section \ref{exp}, it is usually
However, it is usually better to table every predicate whose answers
have multiple explanations and are going to be reused often.

\subsection{PITA Library Functions for the General Probabilistic Case}
\label{gpc}
In the case of general probabilistic programs, the data structure for representing probabilistic information is a Binary Decision Diagram.
With such a data structure, we can represent the explanations for the queries in a form in which they are mutually exclusive and so the computation of the probability can be performed by an effective dynamic programming algorithm.

The predicates that manipulate the data structure in this case
manipulate BDDs. In our implementation, these calls provide a Prolog
interface to the functions in the CUDD C library (\url{http://vlsi.colorado.edu/~fabio/CUDD}). 
The predicates for
  interfacing with CUDD are %\footnote{BDDs
%  are represented in CUDD as pointers to their root node.}
%
\begin{itemize}
  \item \textit{init, end}: for allocation and deallocation of a BDD manager, a data structure used to keep track of the memory for storing BDD nodes;
  \item \textit{zero(-B), one(-B), and(+B1, +B2, -B), or(+B1, +B2, -B), not(+B1, -B)}: Boolean operations between BDDs;
  \end{itemize}

%\section{PRISM Simpler Setting}
\section{\algorithmprism}
\label{prism}
%The PRISM system imposes special requirements on the form of the
%program it can handle. If these requirements are not met, the
%probabilities returned by PRISM are not correct.
As discussed in Section~\ref{problp}, general Probabilistic Logic
Programming requires the computation of the probability of DNF
formulas -- a difficult problem.  The PRISM system avoids this
complexity by imposing special requirements on the form of a program
it can correctly evaluate.  These requirements are~\cite{PRISM-manual}

\begin{itemize}
\item the probability of a conjunction $(A,B)$ is
computed as the product of the probabilities of A and B ({\em independence} assumption)
\item
the probability of a disjunction $(A;B)$ is computed as the sum of
the probabilities of A and B
({\em exclusiveness} assumption).
  \end{itemize}
%The program has to be written so that these requirements are met.
%This is not always possible
It is  possible to write  programs so that these requirements are not met.
For example, consider the program
%Not all programs satisfy the two conditions
%  \begin{itemize}
%  \item Coin, Pea plants, Blood type, Growing negated body satisfy both
%  \item Russian roulette satisfies and
%  \item Dice satisfies or
%  \item Path, Growing head, UWCSE does not satisfy any
%  \end{itemize}
$$
\begin{array}{ll}
p \leftarrow a,b. &   a:0.3 \vee b:0.4.       
\end{array}
$$
This program does not satisfy the independence assumption because the conjunction $a,b$ has probability 0, since $a$ and  $b$ are never true in the same world.
\algorithmprob{} correctly gives probability 0 for $p$ while PRISM returns probability $0.12$. In this case the conjunction $(a,b)$ is inconsistent and, while \algorithmprob{} automatically recognizes it, the inconsistency must be detected and the clause removed for PRISM to return the correct probability.
The following example also does not  satisfy the independence assumption
because $a$ and $b$ both depend on $c$. \algorithmprob{} returns $0.2$ for the probability of $q$  while PRISM returns $0.04$.
$$
\begin{array}{llll}
q\leftarrow a,b.  &  a\leftarrow c. &   b\leftarrow c.&   c:0.2.
\end{array}
$$

As a final example, the following program violates the exclusiveness
assumption as the two clauses for the ground atom $q$ have
non-exclusive bodies
$$
\begin{array}{llll}
q\leftarrow a.   &   q\leftarrow b. &   a:0.2.  &  b:0.4.
\end{array}
$$
% maybe mention here the complexity gain from PRISM?
% FR: I don't know it precisely, I think we could say that evaluating a PRISM program is polinomial as it can be converted to a normal logic program under the WFS, while for LPADs you have to use BDD which have exponential cost.
%The limitations considered by PRISM simplify considerably the
%computation since we do not have to take into account anymore the
%dependencies between the explanations of differen subgoals.
These restrictions required by PRISM simplify considerably the
computation since we can now ignore 
%do not have to take into account anymore 
the dependencies between the explanations of different subgoals.

%PITA can be optimized for this simpler setting. The program
%transformation is simplified and we report here the case for a
%disjucntive clause in order to illustrate it
PITA can be optimized for PRISM-style programs by simplifying the
program transformation it uses, and by implementing simpler library
functions.  
%We illustrate the transformation for the case of a
%disjunctive clause.
%
The  clause
$C_r=H_1:\alpha_1\vee \ldots\vee H_n:\alpha_n\leftarrow L_1,\ldots,L_m$
\noindent
is transformed into the set of clauses $PITA^P(C_r)$
$$\begin{array}{ll} 
PITA^P(C_r,1)=PITA_H(H_1)\leftarrow&one(DD_0),\\
&PITA_B(L_1),and(DD_0,D_1,DD_1),\ldots,\\
&PITA_B(L_m),and(DD_{m-1},D_m,DD_m),\\
&equality([\alpha_1,\ldots,\alpha_n],1,DD),\\
&and(DD_m,DD,D).\\
\ldots\\
PITA^P(C_r,n)=PITA_H(H_n)\leftarrow&one(DD_0),\\
&PITA_B(L_1),and(DD_0,D_1,DD_1),\ldots,\\
&PITA_B(L_m),and(DD_{m-1},D_m,DD_m),\\
&equality([\alpha_1,\ldots,\alpha_n],n,DD),\\
&and(DD_m,DD,D).
\end{array}$$
\noindent
The auxiliary data structure stored in the extra subgoal argument is
no longer a BDD, but simply a real number that represents the
probability of a ground instantiation of that subgoal.  The library
functions are now simple Prolog predicates.
%The auxiliary predicates that are added by the transformation are no longer calls to an external BDD library but are simply defined as 
$$\begin{array}{ll} 
equality(Probs,N,P)\leftarrow nth(N,Probs,P).\\
or(A,B,C)\leftarrow C\ is\  A+B. &   and(A,B,C)\leftarrow C \  is \  A*B.\\
not(P,P1)\leftarrow P1\  is \  1-P.\\
zero(0.0).       &        one(1.0).\\
ret\_prob(P,P).
\end{array}$$
We call the resulting algorithm \algorithmprism{}. 

\begin{figure}[ltb]
\subfigure
	[\algorithmprism{} and PRISM on random sequences.\label{phmm-rand}]{\includegraphics[width=.44\textwidth]{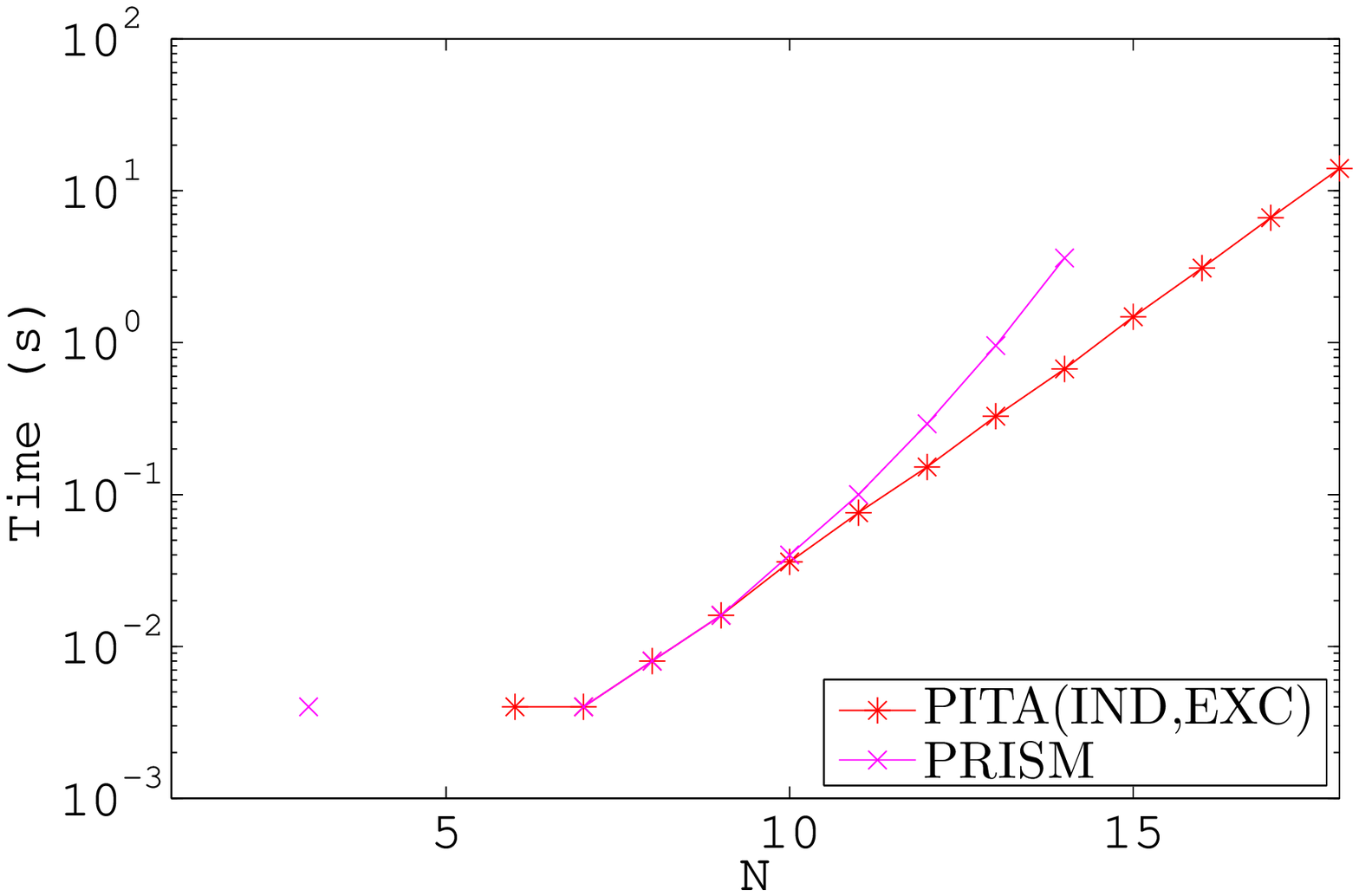}}
\hspace{0.2cm}
\subfigure
	[\algorithmprism{} and PRISM on repeated sequences.\label{phmm-rep}]	{\includegraphics[width=.44\textwidth]{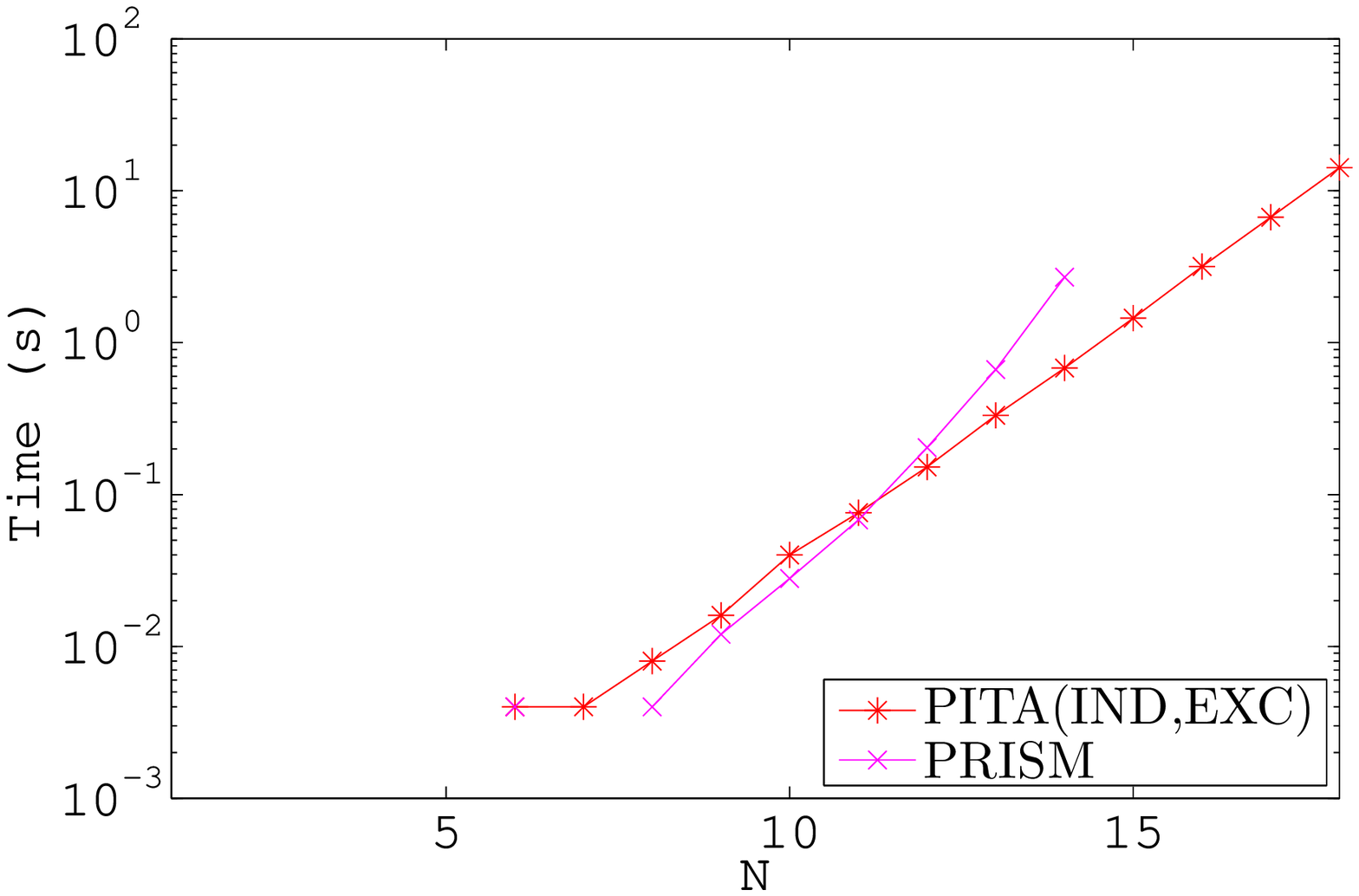}}
\caption{Times for computing $P(hmm(<seq>))$ as a function of sequence
  length. Missing points at the beginning of the $X$-axis correspond
  to a time smaller than $10^{-6}$ seconds, missing points at the end
  of the $X$-axis correspond to a memory error.  The experiments were
  performed on a Core 2 Duo E6550 (2333 MHz) processor.}
\label{phmm}
\end{figure}

An example of a program satisfying the PRISM requirements encodes a
Hidden Markov Model (HMM), a graphical model with a sequence of
unobserved state variables, a sequence of observed output variables,
and where each state variable depends only on its preceding state. 
HMMs have a wide range of applications, including the modeling of DNA sequences. The following program, taken from \cite{DBLP:conf/iclp/ChristiansenG09} models DNA sequences using three states:
% DNA using three states:
%\begin{verbatim}
%hmm(O):-hmm1(_,O).
%hmm1(S,O):-hmm(q1,[],S,O).
%hmm(end,S,S,[]).
%hmm(Q,S0,S,[L|O]):- Q\= end, next_state(Q,Q1,S0), letter(Q,L,S0),
%  hmm(Q1,[Q|S0],S,O).
%next_state(q1,q1,_S):1/3;next_state(q1,q2_,_S):1/3;
%  next_state(q1,end,_S):1/3.
%next_state(q2,q1,_S):1/3;next_state(q2,q2,_S):1/3;
%  next_state(q2,end,_S):1/3.
%letter(q1,a,_S):0.25;letter(q1,c,_S):0.25;
%  letter(q1,g,_S):0.25;letter(q1,t,_S):0.25.
%letter(q2,a,_S):0.25;letter(q2,c,_S):0.25;
%  letter(q2,g,_S):0.25;letter(q2,t,_S):0.25.
%\end{verbatim}
{\em \begin{tabbing}
fooooo\==foooooooooooooooooo\=ooooooooooooooooo\=ooooooooooooo\=\kill
hmm(O) $\leftarrow$ hmm1(\_,O).\\
hmm1(S,O)$\leftarrow$ hmm(q1,[],S,O).\\
hmm(end,S,S,[]).\\
hmm(Q,S0,S,[L$|$O])$\leftarrow$ Q $\backslash=$  end, succ(Q,Q1,S0), out(Q,L,S0),\\
\>  hmm(Q1,[Q$|$S0],S,O).\\
succ(q1,q1,\_S):1/3 $\vee$ succ(q1,q2,\_S):1/3 $\vee$ succ(q1,end,\_S):1/3.\\
succ(q2,q1,\_S):1/3 $\vee$ succ(q2,q2,\_S):1/3 $\vee$ succ(q2,end,\_S):1/3.\\
out(q1,a,\_S):1/4 $\vee$ out(q1,c,\_S):1/4 $\vee$ out(q1,g,\_S):1/4 $\vee$ out(q1,t,\_S):1/4.\\
out(q2,a,\_S):1/4 $\vee$ out(q2,c,\_S):1/4 $\vee$ out(q2,g,\_S):1/4 $\vee$ out(q2,t,\_S):1/4.
\end{tabbing}}
In order to investigate the relative performances of \algorithmprism{}
and PRISM, we computed the execution time of queries to {\tt hmm/1} for increasing lengths of the
output sequence.  Sequences used in Figure~\ref{phmm-rand} are
randomly generated, while those in Figure~\ref{phmm-rep}
%\footnote{The experiments
%  were performed on a  Core 2 Duo E6550
%  (2333 MHz) processor.  } 
are repetitions of the sequence {\tt a,c,g,t}. (Version 2.0 of Prism
was used in all the experiments.)  In both cases, the costs for both
algorithms grow exponentially. Times for both systems are close for
$N$ up to 11; however beyond $N=12$, \algorithmprism{} begins to scale
somewhat better than Prism, answering queries through $N=18$ while
Prism can answer queries only through $N=14$.  Beyond those numbers, both
systems throw memory errors.
% was here
%\begin{figure}
%\begin{center}
%\includegraphics[scale=0.32,draft=false]{time_hmm_pita_prism}
%\end{center}
%\caption{Time for computing $P(hmm([a,\ldots,a])$ as a function of sequence length. Missing points at the beginning of the $X$-axis correspond to a time smaller than $10^{-6}$ seconds, missing points at the end of the $X$-axis correspond to a memory error.}
%\end{figure}

%-----------------------------------------
%-----------------------------------------
%\section{Further Optimizations}
%\label{cg}
\cite{DBLP:conf/iclp/ChristiansenG09} proposed a technique for speeding up query answering by removing
{\em non-discriminating arguments}. These are arguments that play no role in determining the control flow of a logic program with
respect to goals satisfying given mode and sharing restrictions.
The computation trees of the resulting program are isomorphic to those of the original program and 
the results of the original program can be reconstructed from a
trace of the transformed program. 
The authors show that the removal of non-discriminating arguments is very
useful with tabling because the calls to a tabled predicate differing
only in the non-discriminating arguments will merge into a single
table that is much smaller and has a higher chance of reuse.
% Thus the table will be much smaller and there will be a larger
%chance that the current call has a match in the table.
%
After removing non-discriminating arguments, the HMM program above becomes
{\em \begin{tabbing}
fooooo\==foooooooooooooooooo\=oooooooooooo\=ooooooooooooo\=\kill
\> hmm(O)$\leftarrow$ hmm(q1,O).\\
\> hmm(end,[]).\\
\> hmm(Q,[L$|$O])$\leftarrow$ Q $\backslash=$ end, succ(Q,Q1,S0),out(Q,L,S0),hmm(Q1,O).
\end{tabbing}}
%$$\begin{array}{l} 
%hmm(O)\leftarrow hmm(q1,O).\\
%hmm(end,[]).\\
%hmm(Q,[L|O]):- Q\backslash= end, succ(Q,Q1,S0),out(Q,L,S0),\\
%\ \   hmm(Q1,O).
%\end{array}$$
\noindent
plus the clauses defining $succ/2$ and $out/2$.

%------------------------------------------------------------------
\begin{figure}[ltb]
\subfigure
%	[\algorithmprism{} and PRISM.\label{phmmhc-rand}]{\includegraphics[width=.47\textwidth]{time_hmmhc_pita_prism}}
[\algorithmprism{} and PRISM on random sequences.\label{phmmhc-rand}]{\includegraphics[width=.44\textwidth]{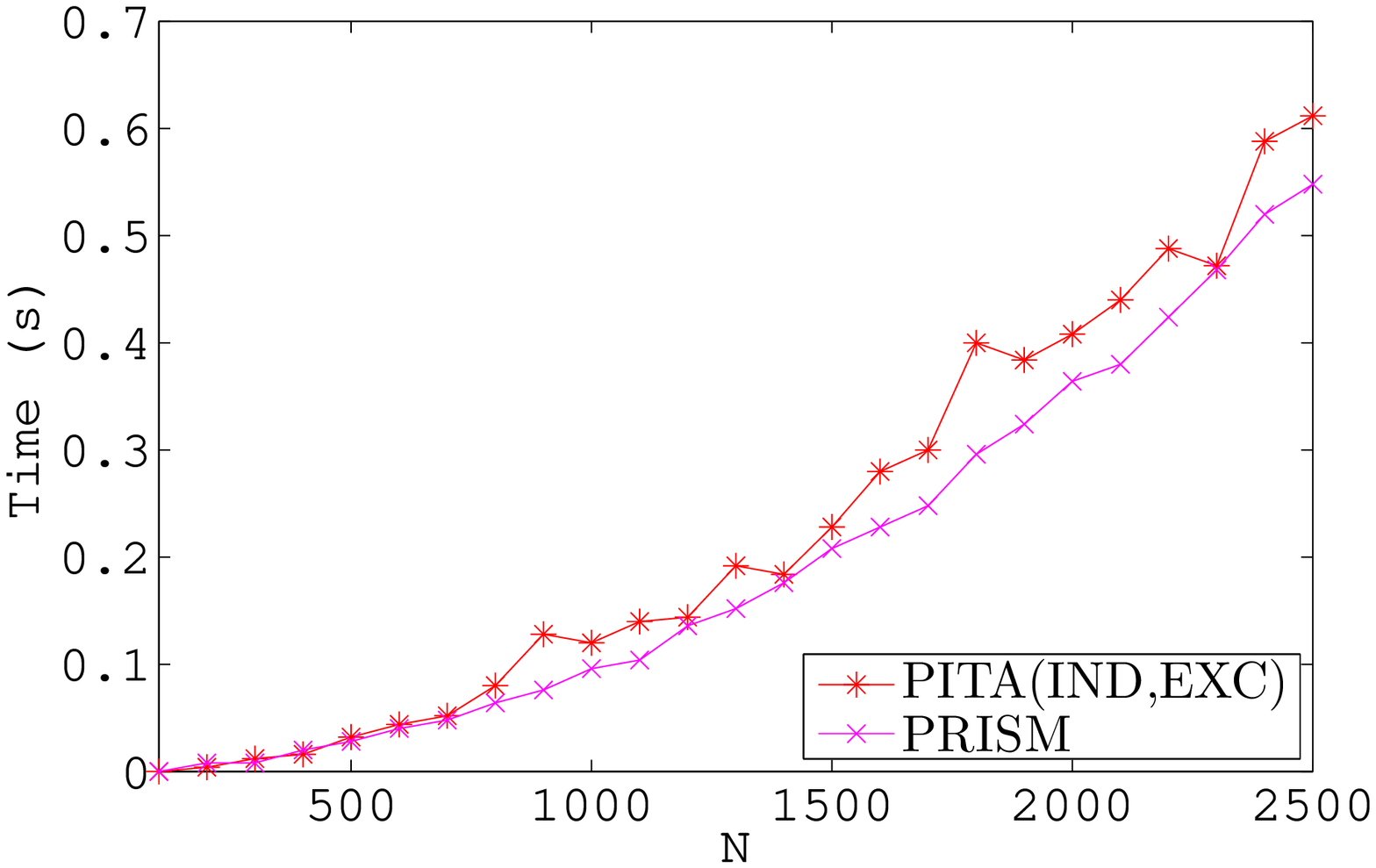}}
\hspace{0.2cm}
\subfigure
	[\algorithmprism{} and PRISM on repeated sequences.\label{phmmhc-rep}]	{\includegraphics[width=.44\textwidth]{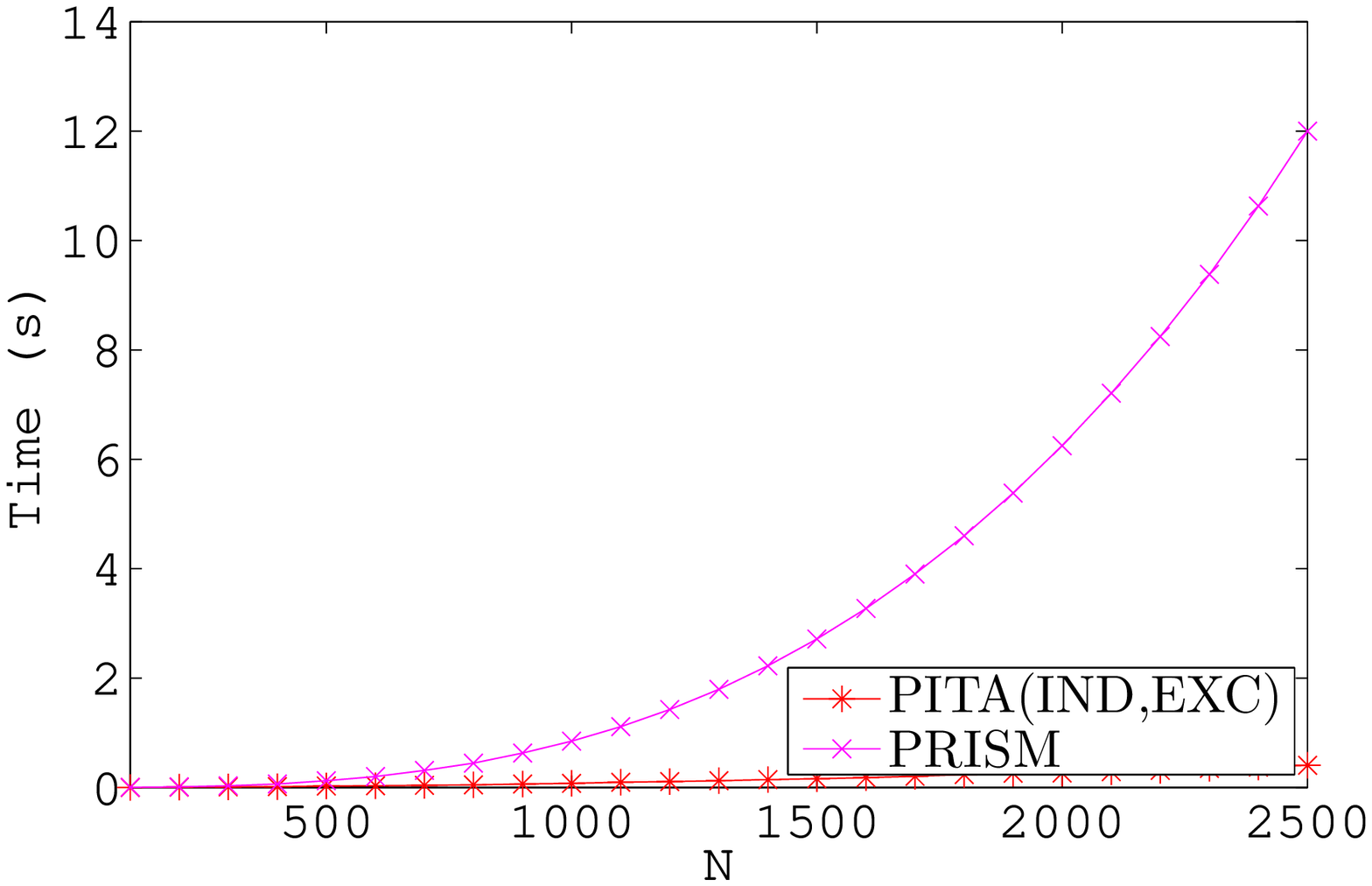}}
\caption{Times for computing $P(hmm(<seq>)$ as a function of sequence
  length (reduced program with non-discriminating arguments removed).
  The experiments were performed on a Core 2 Duo E6550 (2333 MHz)
  processor.}  
%\vspace{-0.8cm}
\end{figure}
%------------------------------------------------------------------

Figures \ref{phmmhc-rand} and \ref{phmmhc-rep}
%\footnote{The experiments
%  were performed on a  Core 2 Duo E6550
%  (2333 MHz) processor.} 
show the computation time for \algorithmprism{} and PRISM on the
reduced HMM program as a function of the sequence length for randomly
generated and repeating sequences.
For random sequences, \algorithmprism{} and Prism are
competitive, with Prism slightly faster; however for the repeating
sequences \algorithmprism{} is much faster, and in fact scales well up
to input sequences of length ${\cal O}(10^5)$.  The reason for the
scalability of \algorithmprism{} on repeated sequences is apparently
due to XSB's use of trie-based tables, which allows good indexing 
and space sharing for repeating subsequences.  The tabling of Prism,
which is based on hash tables, loses discrimination in this case.
% I think we're pretty close to linear (?)  That is almost how
% the graphs look.
%Note that neither system is linear in the sequence length (cf.
%\cite{DBLP:conf/iclp/ChristiansenG09}).  The non-linearity appears to
%be due to extra memory managerment operations needed to table lists of
%larger sizes when the sequence size is increased..

\comment{ Figure \ref{phmmhc}
  shows the computation time for \algorithmprism{} and PRISM on the
  reduced program as a function of the sequence length, while Figure
  \ref{phmmhcpita} shows the behavior of \algorithmprism{} on extended
  lengths. \algorithmprism{} is faster for all values of the sequence
  length and scales to very large values of the length
for the query $P(hmm([a,\ldots,a])$.  When the input sequence is a
random list of letters, \algorithmprism{} also scales well to lengths
of about 10000.  We believe the reason for the scalability of
\algorithmprism{} is that XSB's use of trie-based tables is efficient
for storing sublists, especially when the lists have repeating
elements.  
We also
tried PRISM with a sequence length of 20000 but obtained a memory
error.
}

\paragraph*{Computing the Viterbi Path.}
%{\em Computing the Viterbi Path.}
%In the previous section we concentrated on computing the probability
%of the goal \verb|hmm([a,...,a])| for increasing sequence length. 
In HMMs, it is common to look for the sequence of state values that
most likely gave the output sequence, also called the \emph{Viterbi
  path}, while the probability of this sequence of states is called
the \emph{Viterbi probability}. This is equivalent to finding the most
probable explanation for the goal.

The Viterbi path and probability are computed by PRISM with the $\mathit{viterbif}/3$ predicate but can be computed also by \algorithmprism{} by modifying it so that the probability data structure includes not only the highest probability of the subgoal but also the most probable explanation for the subgoal. In this case the support predicates are modified as follows:
{\em \begin{tabbing}
fooooo\==foooooooooooooooooo\=ooooooooooooooooo\=ooooooooooooo\=\kill
\> equality(R,S,Probs,N,e([(R,S,N)],P))$\leftarrow$ nth(N,Probs,P).\\
\> or(e(E1,P1),e(\_E2,P2),e(E1,P1))$\leftarrow   P1 >=P2,!.$\\
\> or(e(\_E1,\_P1),e(E2,P2),e(E2,P2)).\\
\> and(e(E1,P1),e(E2,P2),e(E3,P3))$\leftarrow$    P3\ is\  P1*P2,append(E1,E2,E3).\\
\> zero(e(null,0)).                  
\> one(e([],1)).
\> ret\_prob(B,B).
\end{tabbing}}
\noindent
In this way we obtain \algorithmvitind{}, which is also sound if the
exclusiveness assumption does not hold.

\begin{figure}[hbt]
\subfigure
	[\algorithmvitind{} and PRISM on random sequences.\label{phmmvit-rand}]{\includegraphics[width=.44\textwidth]{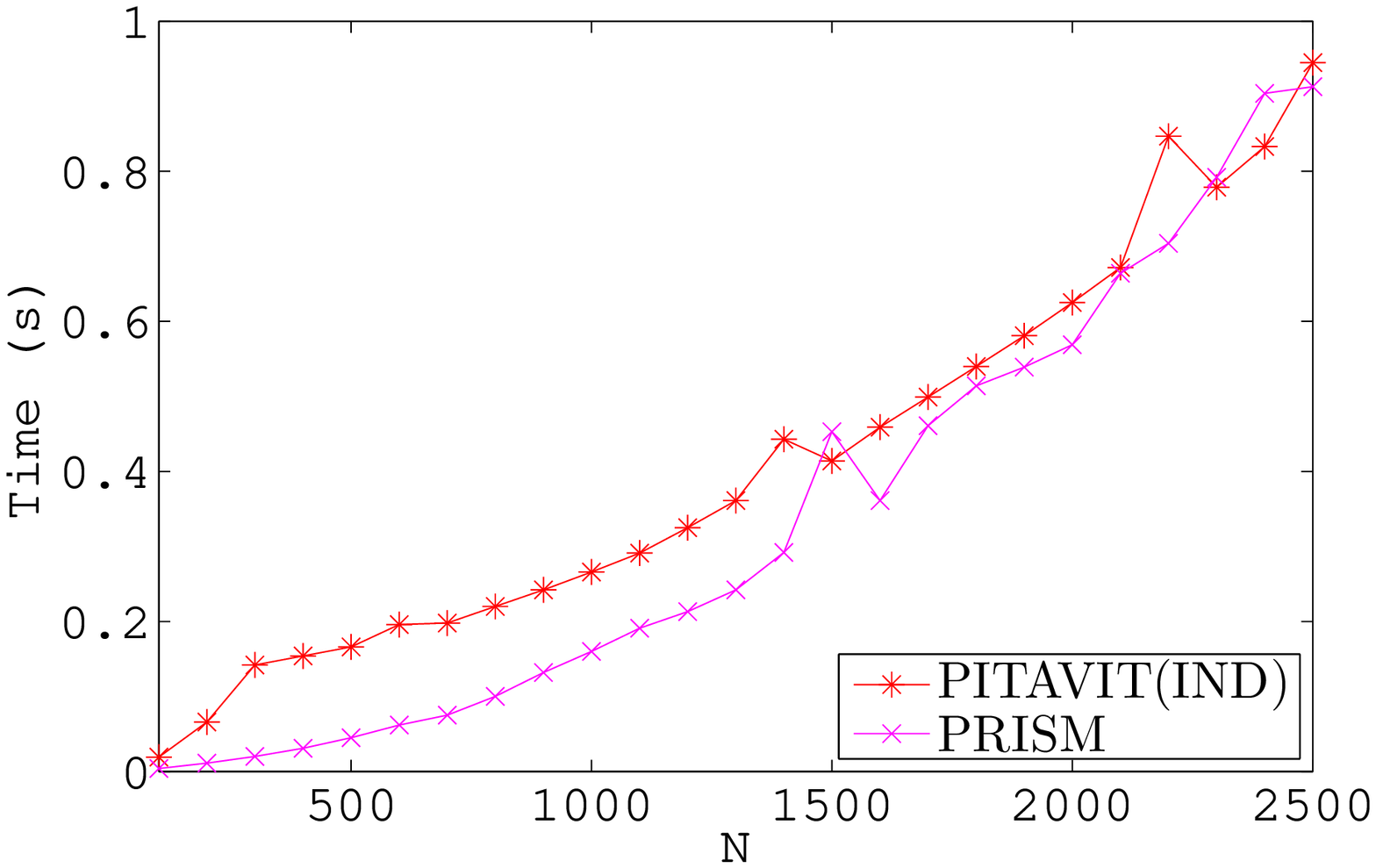}}
\hspace{0.2cm}
\subfigure
	[\algorithmvitind{} and PRISM on repeated sequences.\label{phmmvit-rep}]{\includegraphics[width=.44\textwidth]{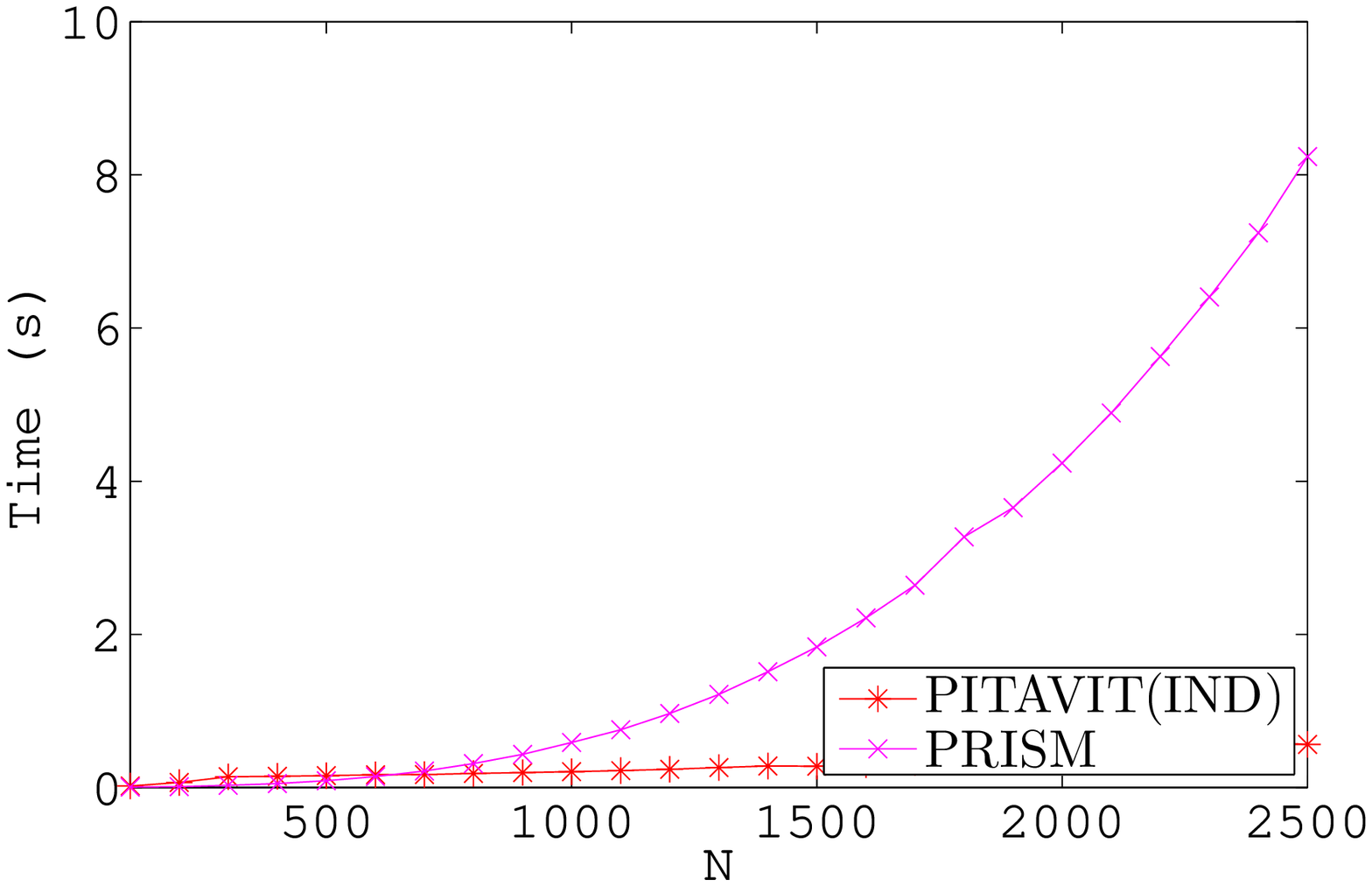}}
\caption{Times for computing the Viterbi path and probability of
  $hmm(<seq>)$ as a function of sequence length (reduced program with
  non-discriminating arguments removed).  The timings were taken on an
  Intel Core i5 (2.53 GHz) processor.}
\end{figure}

Figures \ref{phmmvit-rand} and \ref{phmmvit-rep}
%\footnote{The timings 
%  were taken on an Intel Core i5 (2.53 GHz) processor.} 
show times for
\algorithmvitind{} and PRISM to compute Viterbi paths and
probabilities on the reduced HMM program.  \algorithmvitind{} is slower than PRISM for short random
sequences and roughly the same on long sequences. On repeated sequences it is much more scalable.

%Figures \ref{phmmvit_rand} and \ref{phmmvit_rep}
%as a function of sequence length for \algorithmprism{} and PRISM.
%Figure \ref{phmmvitpita} shows the performance of \algorithmprism{}
%for larger lengths. \algorithmprism{} shows a speedup similar to the
%previous case. We also tried PRISM with a sequence length of 20000 but
%obtained a memory error.  Note that the time for both systems is not
%linear in the sequence length as is should be according to the
%theoretical considerations of \cite{DBLP:conf/iclp/ChristiansenG09}.
%The non linearity is probably due to bookkeeping operations.  
%The non-linearity appears to be due to memory managerment operations
%need to table lists of larger sizes when the input sequence size is increased.

%------------------------------------------------------------
%------------------------------------------------------------

{\em Counting Explanations}
%%%%%%%%%%%%%%%%%%%%%%%%%%%%%%%%%%%%%%
%I don't quite get this?
% FR: for program
%p:- a,b.             
%a:0.3 ; b:0.4.       
% the explanations for a and b are incompatible but PITA says there is 1 explanation, while there is 0.
\algorithmprism{} can be used to count explanations for goals with a slight modification when 
explanations for different goals are not incompatible.
To obtain PITA(COUNT), the only auxiliary predicate to be modified is $equality/3$:
$\begin{array}{ll} 
equality(\_Probs,\_N,1).
\end{array}$

%%%%%%%%%%%%%%%%%%%%%%%%%%%%%%%%%%%%%%

\section{Application to Possibilistic Logic Programming}
\label{appposslp}
PITA also can be used to perform inference in Possibilistic Logic
Programming where a program is composed only of clauses of the form
$H:\alpha\leftarrow B_1,\ldots,B_n$ which we interpret as
possibilistic clauses of the form $(H\leftarrow
B_1,\ldots,B_n,\alpha)$.  For space reasons we do not discuss negation
here, however the publicly available version of PITA computes
possibilistic programs that are left-to-right dynamically stratified
(Section~\ref{tabling}) according to the semantics of~\cite{BSCV10}.

The transformation $PITA^P$ used for the PRISM optimization can be used unchanged provided the support predicates are defined as
{\em \begin{tabbing}
fooooo\==foooooooooooooooooo\=ooooooooooooooooo\=ooooooooooooo\=\kill
\> equality([P,\_P0],\_N,P). \\
\> or(A,B,C)$\leftarrow$ C\ is\ max(A,B). \> \>
   and(A,B,C)$\leftarrow$ C is min(A,B).\\
\> zero(0.0). \> one(1.0). \> ret\_prob(P,P). 
\end{tabbing}}
\comment{
$$\begin{array}{ll} 
equality([P,\_P0],\_N,P).\\
or(A,B,C)\leftarrow C\ is\ max(A,B).\\
and(A,B,C)\leftarrow C is min(A,B).\\
zero(0.0).\\
one(1.0).\\
ret\_prob(P,P).\\
\end{array}$$}
\noindent
We obtain in this way \algorithmposs{}. The input list of the
$equality/3$ predicate contains two numbers because we used the same
preprocessing code as for LPADs. Specializing the transformation for
possibilistic logic programs would remove the need for the
$equality/3$ predicate.  

To experiment with \algorithmposs{}, we consider the networks of
biological concepts of \cite{DBLP:conf/ijcai/RaedtKT07} and the
definition of \emph{path/2} of Example \ref{poss-path}. In these
networks the nodes encode biological entities and the edges
conceptual relations among them. In each program the edges are
associated to a real number. The programs have been sampled from a
very large graph and contain 200, 400, $\ldots$, 10000 edges.
Sampling was repeated ten times, to obtain ten series of programs of
increasing size. In each program we query the possibility that the two
genes HGNC\_620 and HGNC\_983 are related.

%To get an idea of the complexity of the problem, 
We use PITA(COUNT) to compute the number of explanations for the query
in the first series of programs.  In this problem, an explanation
is a path from source to target that does not contain loops. In fact,
paths with loops are subsumed by paths without loops so they do not
contribute to the overall probability.
%Since the \verb|or/3| operation of PITA(COUNT) is not idempotent, we used the definition of path of
% \cite{DBLP:conf/iclp/KimmigCRDR08} that performs loop checking
% explicitly by keeping the list of visited nodes.
Table \ref{npaths} shows the number of paths for the networks in series 1 for which the computation terminated in 24 hours.
\begin{table}[hbt]
\label{npaths}
\caption{Number of paths.}
%\begin{minipage}{\textwidth}
{\footnotesize
\begin{tabular}{|l|r|r|r|r|r|r|}
\hline
\hline
Edges& 200& 400&600&800&1000&1200\\\hline
Explanations&10& 42 &  380&1,280&3,480&612,140\\\hline\hline
\end{tabular}}
%\end{minipage}
\end{table}
As you can see, the number of paths grows very fast.
%%
%The definition of \verb|path| implies that these are also the counts of the number of distinct paths from source to target that do not contain loops

\begin{figure}
\subfigure
	[Average time.\label{posslog}]{\includegraphics[width=.44\textwidth]{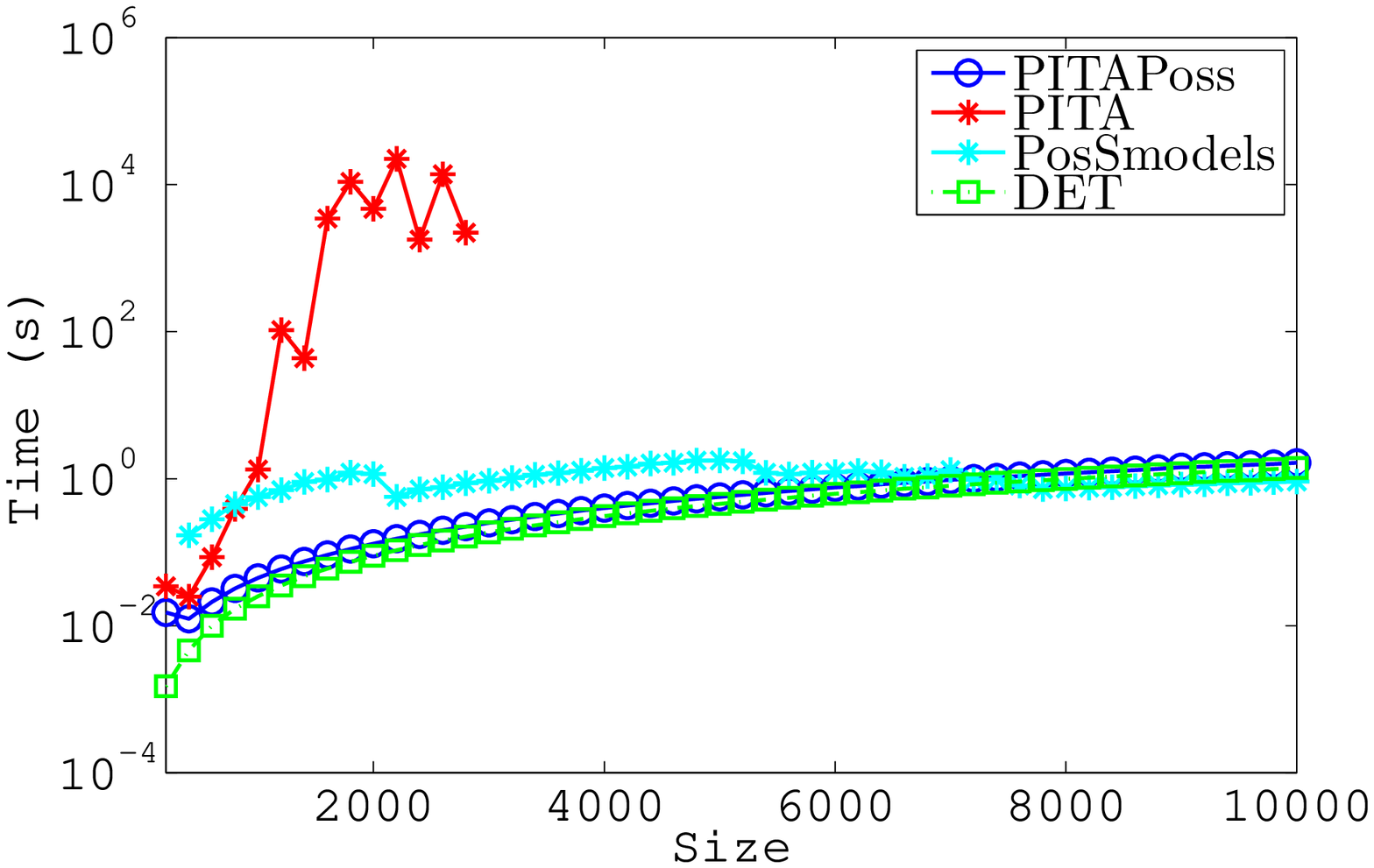}}
\hspace{0.2cm}
\subfigure
	[Number of solved graphs.\label{posssolved}]	{\includegraphics[width=.44\textwidth]{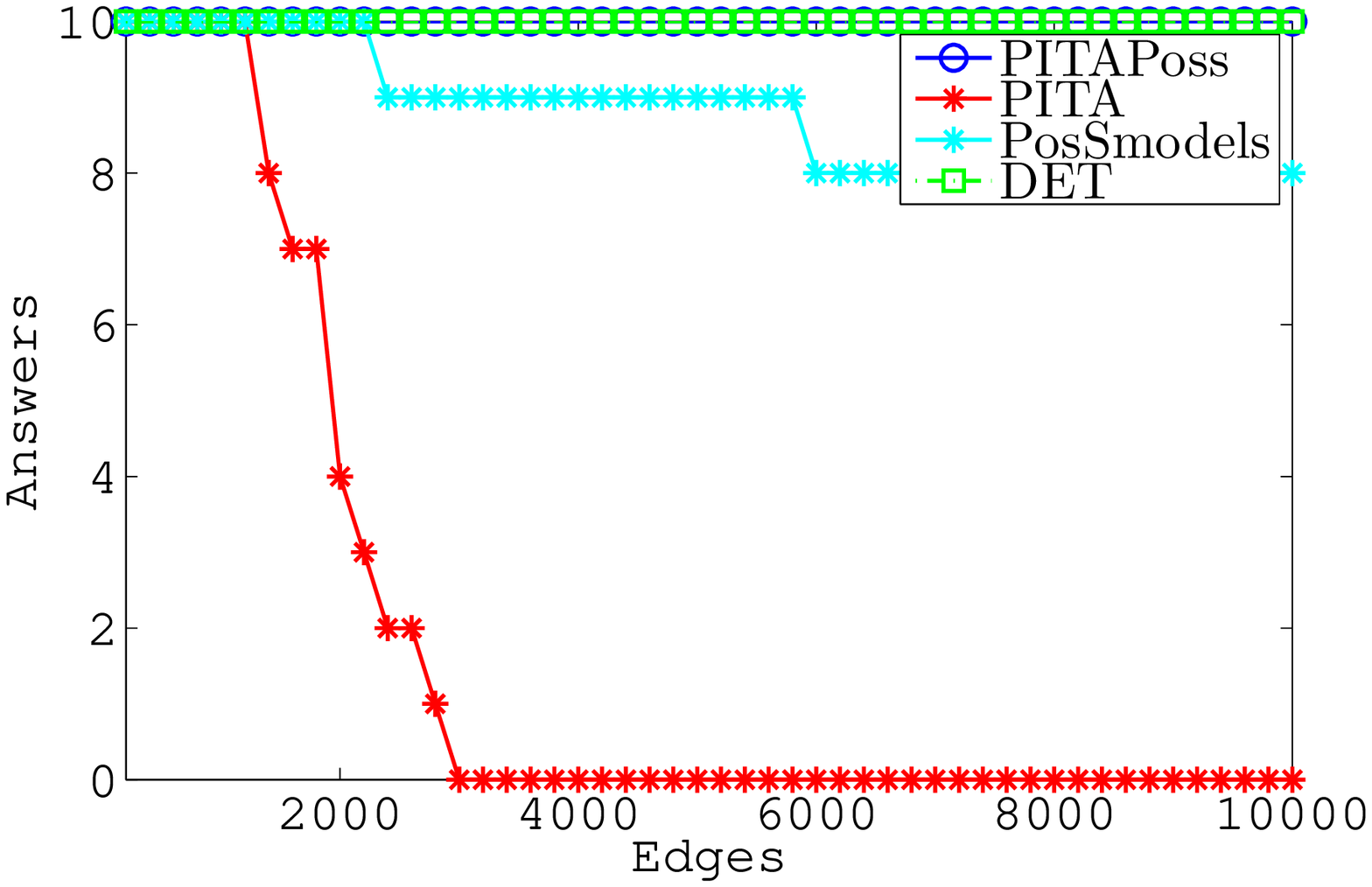}}
	
%\begin{center}
%\includegraphics[scale=0.32,draft=false]{time_kimmig_pita_pitaposs}
%\end{center}

\caption{Results of the experiments on the biological networks.  The
  experiments were performed on an Intel Core 2 Duo E6550 (2333 MHz)
  processor and 4 GB of RAM.} 
\label{biomine}
\vspace{-0.18in}
\end{figure}

Figure \ref{posslog}
% \footnote{The experiments
%  were performed on an Intel Core 2 Duo E6550 (2333 MHz) processor and
%  4 GB of RAM.} 
shows  the average over the ten series of the execution
time for computing the possibility of
\emph{path('HGNC\_620','HGNC\_983')} as a function of the number of
edges. Figure \ref{posssolved} shows the number of graphs solved for each graph size. These figures also contain data for
\algorithmprob{}, 
for the equivalent deterministic program
(i.e. computing whether there is a path between nodes) and for the
system posSmodels \cite{DBLP:journals/amai/NicolasGSL06}\footnote{For \algorithmprob{}, we used the definition of path of
  \cite{DBLP:conf/iclp/KimmigCRDR08} because it gave smaller
  timings. \algorithmprism{} was not tested because this problem does
  not satisfy the independence and exclusiveness requirements}.  As
these figures show, computing the possibility is much easier than
computing the general probability, which must solve the disjoint sum
problem to obtain answers.  With respect to the posSmodels system,
\algorithmposs{} is faster for smaller graphs and slower for larger
ones, but the averages of posSmodels have been computed on less graphs
since on some it gave a lack of memory error.

\vspace{-0.12in}
\input{conc}

\textit{Acknowledgements} The authors thank Henning
Christiansen for his help in validating the experimental results that
use removal of non-discriminating arguments.
The work of the first author has been partially supported by the Camera di Commercio, Industria, Artigianato e Agricoltura di Ferrara, under the project titled "Image Processing and Artificial Vision for Image Classifications in Industrial Applications".
%\vspace{-0.5cm}
\bibliographystyle{acmtrans}
\bibliography{strings,bib,lpadlocal}

\end{document}

%% file: abstract.tex
\begin{abstract}
Many real world domains require the representation of a measure of
uncertainty.  The most common such representation is probability, and
the combination of probability with logic programs has given rise to
the field of Probabilistic Logic Programming (PLP), leading to
languages such as the Independent Choice Logic, Logic Programs with
Annotated Disjunctions (LPADs), Problog, PRISM and others. These languages
share a similar distribution semantics, and methods have been devised
to translate programs between these languages. 
The complexity of computing the probability of queries to these
general PLP programs is very high due to the need to combine the
probabilities of explanations that may not be exclusive.  As one
alternative, the PRISM system reduces the complexity of query
answering by restricting the form of programs it can evaluate.  As an
entirely different alternative, Possibilistic Logic Programs adopt a
simpler metric of uncertainty than probability.

Each of these approaches -- general PLP, restricted PLP, and
Possibilistic Logic Programming -- can be useful in different domains
depending on the form of uncertainty to be represented, on the form of
programs needed to model problems, and on the scale of the problems to
be solved.  In this paper, we show how the PITA system, which
originally supported the general PLP language of LPADs, can also
efficiently support restricted PLP and Possibilistic Logic Programs.
PITA relies on tabling with answer subsumption and consists of a
transformation along with an API for library functions that interface
with answer subsumption.  We show that, by adapting its transformation
and library functions, PITA can be parameterized to \algorithmprism{}
which supports the restricted PLP of PRISM, including optimizations
that reduce non-discriminating arguments and the computation of
Viterbi paths.  Furthermore, we show PITA to be competitive with PRISM
for complex queries to Hidden Markov Model examples, and sometimes
much faster.
%Furthermore, we show PITA to be faster and more scalable than
%PRISM for complex queries to Hidden Markov Model examples.  
We further show how PITA can be parameterized to PITA(COUNT) which
computes the number of different explanations for a subgoal, and to
PITA(POSS) which scalably implements Possibilistic Logic Programming.
PITA is a supported package in version 3.3 of XSB.
\end{abstract}

%Furthermore, we show PITA to be faster and more scalable than
%PRISM for complex queries to Hidden Markov Model examples.  
%
%Furthermore, we show PITA competitive with PRISM for complex queries
%to Hidden Markov Model examples, and sometimes much faster.

%% file: tab.tex
\section{Tabling and Answer Subsumption}
\label{tabling}
The idea behind tabling is to maintain in a table both subgoals
encountered in a query evaluation and answers to these subgoals.  If a
subgoal is encountered more than once, the evaluation reuses
information from the table rather than re-performing resolution
against program clauses.  Although the idea is simple, it has
important consequences.  First, tabling ensures termination for a wide
class of programs, and it is often easier to reason about termination
with programs using tabling than with basic Prolog.  Second, tabling can be used to
evaluate programs with negation according to the WFS.  Third, for
queries to wide classes of programs, such as datalog programs with
negation, tabling can achieve the optimal complexity for query
evaluation.  And finally, tabling integrates closely with Prolog, so
that Prolog's familiar programming environment can be used, and no
other language is required to build complete systems.  As a result, a
number of Prologs now support tabling including XSB, YAP, B-Prolog,
ALS, and Ciao.  In these systems, a predicate $p/n$ is evaluated using
SLDNF by default: the predicate is made to use tabling by a
declaration such as {\em table p/n} that is added by the user or
compiler.

This paper makes use of a tabling feature called {\em answer
  subsumption}.  Most formulations of tabling add an answer $A$ to a
table for a subgoal $S$ only if $A$ is a not a variant (as a term) of
any other answer for $S$.  However, in many applications it may be
useful to order answers according to a partial order or (upper
semi-)lattice.  As an example, consider the case of a lattice on the
second argument of a binary predicate $p$. Answer subsumption may be
specified by means of a declaration such as {\em table p(\_,join/3 -
  bottom/1)}
%
%\begin{verbatim}
%:- table arc(_,_,or/3 - zero/1)).
%\end{verbatim}
%
%which indicates that if a table contains an answer
%$arc(Arg_1,Arg_2,Arg_{1,3})$, and a new answer
%$arc(Arg_1,Arg_2,Arg_{2,3})$ is derived, then
%$arc(Arg_{1},Arg_{2},Arg_{1,3})$ is replaced by
%$arc(Arg_{1},Arg_{2},or(Arg_{1,3},$ $Arg_{2,3}))$ 
where $bottom/1$ returns the bottom element of the lattice and $join/3$ is the
join operation of the lattice. Thus if a table had an answer
$p(a,d_1)$ and a new answer $p(a,d_2)$ were derived, the answer
$p(a,d_1)$ would be replaced by $p(a,d_3)$, where $d_3$ is obtained by
calling $join(d_1,d_2,d_3)$. In the PITA algorithm for LPADs presented
in Section \ref{algorithm}, the last argument of atoms is used to
store explanations for the atom in the form of BDDs and the $join/3$
operation is the logical disjunction of two explanations\footnote{The
  logical disjunction $d_3$ can be seen as subsuming $d_1$ and $d_2$
  over the partial order af implication defined on propositional
  formulas that represent explanations.}; under the simplifying
assumptions of \algorithmprism{} $or/3$ is simple addition; while for
possibilistic logic $or/3$ takes the maximum of its input arguments.
Answer subsumption over arbitrary upper semi-lattices is implemented
in XSB for stratified programs~\cite{Swif99a}; in addition, the
mode-directed tabling of B-Prolog~(cf. \cite{Zhou11}) can also be seen
as a form of answer subsumption.

%\cite{RigSwi10-CILC10-NC} indicates classes of LPADs
%(cf. Section~\ref{problp}) for which tabling with answer subsumption
%is guarenteed to terminate.  The tabling systems considered there use
%a technique called {\em term-size abstraction}, essentially a sound
%method of abstraction for subgoals containing terms above a certain
%size.  
For function-free programs, the tabling used by the PITA system
terminates correctly for left-to-right dynamically stratified LPADs.
However, we note that the termination results
of~\cite{RigSwi10-CILC10-NC} and PITA itself both apply to a much
larger class of well-defined LPADs with function symbols.  As noted in
Section~\ref{problp}, the major probabilistic logic languages defined
under the distribution semantics can be finitely translated into one
another, so that the termination and correctness results for LPADs
extend to other languages: in particular to the restricted PLP
language of Section~\ref{prism}.  In addition the results of
\cite{RigSwi10-CILC10-NC}, which capture termination of general
probabilistic programs that give rise to multiple worlds, directly
apply to the simpler case of Possibilistic Logic Programs, which do
not give rise to multiple worlds.

%% file: conc.tex
\section{Conclusions}
\label{conc}
We have shown how the probabilsitic inference system PITA can be
easily adapted for different settings. In particular, we have
considered programs that respect the independence and exclusion
assumptions that are required by PRISM and show how PITA can be
modified to exploit these assumption. Preliminary results show the
algorithm to be faster than PRISM for complex queries to a naive
encoding of an HMM, while the performance on an optimized encoding
depend on the input data.  Moreover, PITA can be used also for
computing the Viterbi path, i.e., the most probable explanation for a
goal.  Finally, we have shown how PITA can be modified to perform
inference on Possibilistic Logic Programs.

PITA is a supported package in version 3.3 of XSB, and handles
programs that include both negation and function symbols.  Because
PITA consists of a program transformation plus library functions that
implement an API for answer subsumption, the approaches of general
PLP, restricted PLP and Possibilistic Logic Programming can be
combined within a single program.  Thus, if it is known that, say,
predicates in a given module satisfy independence and exclusiveness
assumptions, the module can use \algorithmprism{} and avoid the
expense of BDD maintenance.  Furthermore, simple modifications to PITA
would allow the use of general vs. restricted PLP to be decided on a
predicate basis, possibly supported in the future by an optimizing
compiler that could check exclusiveness of clauses, and independence
of literals within the body of a clause.  This approach is not only
general, but portable.  For Prologs that implement tabling, the
additional effort needed for answer subsumption is relatively small so
that implementations of PITA need not be restricted to XSB.

%In the future we plan to investigate 
%ways of performing 
%approximate inference in order to be able to tackle even larger
%programs. 
Finally, we believe that the techniques presented can be
applied also to Soft Constraint Logic Programming (SCLP)
\cite{DBLP:journals/toplas/BistarelliR01}, as advocated in
\cite{DBLP:journals/entcs/BistarelliMRS07}.
%TLS2 adding sentence -- check if ok
In this case, PITA's API to answer subsumption would interface with a
constraint handling system rather than to BDDs or to simple Prolog
predicates.
In fact, PITA, \algorithmprism{} and \algorithmposs{} can be as seen as
implementing SCLP over the semirings $\langle
\mathcal{P},\vee,\wedge,\mathit{false},true\rangle$, $\langle
        [0,1],+,\times,0,1\rangle$ and $\langle
        [0,1],\max,\min,0,1\rangle$ respectively, where $\mathcal{P}$
        is the set of propositional formulas built over a fixed and
        finite set of propositions.